# New Metrics Between Rational Spectra and their Connection to Optimal Transport

Fredrik Bagge Carlson*    Mandar Chitre

Acoustic Research Laboratory
National University of Singapore

April 20, 2020

**Abstract**

We propose a series of metrics between pairs of signals, linear systems or rational spectra, based on optimal transport and linear-systems theory. The metrics operate on the locations of the poles of rational functions and admit very efficient computation of distances, barycenters, displacement interpolation and projections. We establish the connection to the Wasserstein distance between rational spectra, and demonstrate the use of the metrics in tasks such as signal classification, clustering, detection and approximation.

## 1 Introduction

Measures of distance are fundamental to the fields of optimization, learning and estimation. While we often have an intuitive understanding of the distance between two objects, in some domains, the concept of distance and similarity is harder to grasp.

When comparing two signals, the distance is often calculated in either time domain or in the frequency domain. For the chosen domain, one must also choose a suitable metric under which the distance is calculated. For reasons to be discussed in this introduction, standard choices of metrics, like the $L_2$ norm or the cosine distance, are often poor choices for measuring the distance between two signals, no matter which domain we view the signal in. Recently, the field of optimal transport (OT), and the related methodologies dynamic time warping (DTW) and dynamic frequency warping, have presented promising ways of measuring the distance between two signals in a meaningful way. While these methods allow for interesting things like gradient-based learning and barycenter calculation/clustering, they remain very computationally costly due to the typically large dimension of the representation required.

In this paper, we will draw inspiration from linear-systems theory in order to derive metrics related to optimal transport, between signals, systems and spectra. The key intuition is that any stationary spectrum can be approximated arbitrarily well by a rational function (Stoica and Moses, 2005), i.e., a linear time-invariant transfer function, and that the locations of the zeros and poles of such a function are intimately linked with its spectrum. These poles and zeros will be what we ultimately base our metrics on.

To set the stage and provide some motivating examples for the development of our metrics, consider for instance the comparison between two short audio recordings in the frequency domain. The human ear is insensitive to small shifts of the frequencies present. However, the $L_2$ norm and cosine distances will see two almost orthogonal vectors, causing their distance to be large. When comparing signals in the time-domain, a small shift in phase causes a large discrepancy under most standard signal norms, while our auditory system once again barely notices such a phase change. Optimal transport (Peyré and Cuturi, 2018) has been proposed as a way to measure distance between spectra (Kolouri et al., 2017) that aligns well with what humans consider similar-sounding audio. Computing the optimal transport, or Wasserstein distance of order $p$, $W_p$, between two spectra in a differentiable way is quite expensive, $\mathcal{O}(|\Omega^2|)$. Metrics that admit more efficient calculation are therefore of interest. Furthermore, calculating the barycenter of several spectra under the Wasserstein metric requires solving a large optimization problem. This problem is by no means impossible to solve, in fact many recent works make use of it with great success even on complicated domains in high dimensions (Schmitz et al., 2018), but we nevertheless argue that there is room for improvement in allowing these methods to scale further.

---

*Fredrik.Bagge@nus.edu.sg. Open-source implementations of all algorithms and experiments are available at https://github.com/baggepinnen/SpectralDistances.jl

Within the field of control theory, a distance between LTI systems is of interest for model reduction, optimization of controllers and system identification. Many such distances exists, notably the $H_2$ and $H_\infty$ norms as well as the $\nu$-gap metric, and they all offer unique strengths and drawbacks. The distances considered in this work are not only distances between spectra, but admit and interpretation also as distances between LTI systems.

In summary, we proceed to propose a family of metrics between signals and linear systems in the frequency domain. We will start by reasoning about distances between spectra and about the locations of poles of rational spectra, which will give the reader an intuitive feeling for the proposed metrics. We will give some insights into why these metrics are useful and how they relate to the optimal-transport distance between two spectra. We then demonstrate a number of applications in machine learning and optimization where they indeed are useful, soft and hard clustering of signals, gradient-based learning and acoustic detection.

## 2 Distances Between Signals and Spectra

Before we introduce the optimal-transport based spectral metrics we will spend some time visualizing the objects of interest and the problems associated with traditional signal metrics. Consider two signals $s_1(t)$ and $s_2(t)$, with Fourier spectra $S_1(\omega)$ and $S_2(\omega)$. If, for instance, one signal is identical to the other but with negative sign, and the distance between them is measured by, e.g., the $L_2$ norm in the time domain $\|s_1 - s_2\|_2$, we have the situation depicted in Fig. 1a. In some applications, notably those related to human or animal perception, the negation of a signal does not matter, it sounds the same. The $L_2$ norm, however, sees a very large difference between the two signals. The situation is similar with a small phase shift between $s_1$ and $s_2$, depicted on the right in Fig. 1a.

If we instead consider comparison of signals in the frequency domain, we may have the situation depicted in the top panel of Fig. 1b. Even though the frequency peaks are located nearby each other, the $L_2$ distance between them is the same as if they were much further apart. The $L_2$ distance is essentially measuring the frequency-wise difference between the two spectra, while having no notion of two frequency peaks being close to each other along the frequency axis.

In essence, optimal transport measures the difference between the spectra (technically between measures) horizontally, like in the lower panel of Fig. 1b. With this intuition, it becomes obvious that small shifts in frequencies (phase in the time-domain) imply small distances between signals. The optimal-transport distance is in some settings referred to as the earth-movers distance; If we consider the two objects to be compared as two piles of dirt, the OT distance measures how much effort would be involved in order to rearrange one of the piles to assume the

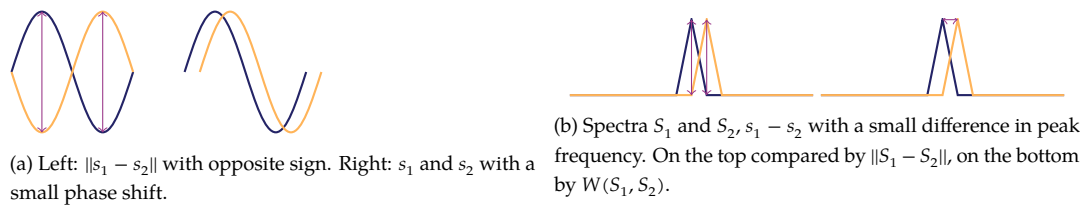

(a) Left: $\|s_1 - s_2\|$ with opposite sign. Right: $s_1$ and $s_2$ with a small phase shift.

(b) Spectra $S_1$ and $S_2$, $s_1 - s_2$ with a small difference in peak frequency. On the top compared by $\|S_1 - S_2\|$, on the bottom by $W(S_1, S_2)$.

Figure 1: The figure illustrates two signals to be compared, in the time domain to the left, and in the frequency domain to the right. The purple arrows indicate roughly how the distance is measured under different metrics.

shape of the other. OT is generally useful when computing distances between two measures of equal mass, such as two probability distributions. The fact that OT, in its standard form, requires the two measures to have equal mass is sometimes an inconvenience. A common heuristic solution is to normalize the measures to integrate to one. This normalization introduces an invariance into the cost function that may or may not be desired depending on the application; For the classification of signals, the signal energy is often of secondary interest, while for control systems, the energy of a spectrum is often of primary concern.

### 2.0.1 Notation

Before we introduce the formalities of optimal transport and rational spectra, we establish some notation. The notation regarding optimal transport largely follows that of Bonneel, Van De Panne, et al. (2011).

$$G(s) = \frac{1}{A(s)} \qquad \text{A stable rational transfer function} \qquad (1)$$

$$\Phi(\omega) = |G(i\omega)|^2 \qquad \text{A rational spectrum} \qquad (2)$$

$$\Omega = \{\omega \in \mathbb{R}\} \qquad \text{A set of frequencies} \qquad (3)$$

$$\gamma \in \Omega \times \Omega \qquad \text{An optimal transport plan} \qquad (4)$$

$$F(\omega) = 2\int_0^\omega \Phi(\varphi)d\varphi \qquad \text{A cumulative spectrum function} \qquad (5)$$

$$F^{-1}(\varepsilon) = \varepsilon \mapsto (\omega : F(\omega) = \varepsilon) \qquad \text{inverse of } F \qquad (6)$$

## 2.1 The Wasserstein and Sinkhorn Distances

Optimal transport is a general way of measuring distance between two measures of equal mass. Recent advances in the ability to compute these distances have led to their rapid adoption as cost functions in machine-learning applications and for computer-graphics applications. These distances are often referred to as the Wasserstein family of distances, and we will use the terms Wasserstein distance and optimal-transport distance interchangeably.

The Wasserstein $p$-distance for $p \geq 1$ is formally given by the solution to the following minimization problem, often referred to as the Kantorovich problem

$$W_p^p(\Phi_1, \Phi_2) = \min_\gamma \int |\omega_1 - \omega_2|^p \, d\gamma(\omega_1, \omega_2) \qquad (7)$$

$$\text{s.t.} \int \gamma(\omega_1, \omega_2)d\omega_2 = \Phi_1(\omega_1), \quad \int \gamma(\omega_1, \omega_2)d\omega_1 = \Phi_2(\omega_2) \qquad (8)$$

For discrete problems, the matrix-vector form looks like

$$W_p^p(\Phi_1, \Phi_2) = \min_\gamma \sum \gamma \odot D \qquad (9)$$

$$\text{s.t.} \sum_j \gamma[i,j] = \Phi_1[i], \quad \sum_i \gamma[i,j] = \Phi_2[j] \qquad (10)$$

where $D \in \mathbb{R}^{|\Omega| \times |\Omega|}$ is the symmetric matrix with all distances $|\omega[i] - \omega[j]|^p$, $\gamma$ is a matrix of same size as $D$, constituting the optimal transport plan and $\odot$ denotes elementwise product.

If an entropy-regularization term is added according to

$$W_{p\lambda}^p(\Phi_1, \Phi_2) = \min_\gamma \sum \gamma \odot D - \frac{1}{\lambda}h(\gamma) \qquad (11)$$

where $h(\gamma)$ is the entropy of $\gamma$, the distance is referred to as the Sinkhorn distance (Cuturi, 2013). This formulation is sometimes beneficial for computational reasons, as instead of solving a linear program, we can now use the more efficient Sinkhorn algorithm. The solution to the regularized form is also more stable with respect to perturbations of the inputs, of particular importance when differentiated.

### 2.1.1 Closed-form solutions to $W$

It is a known fact (Peyré and Cuturi, 2018) that the Wasserstein distance between two continuous measures in one dimension is given by the following distance between inverse cumulative distribution functions

$$W_p^p(\Phi_1, \Phi_2) = \int_0^1 \left| F_1^{-1}(\varepsilon) - F_2^{-1}(\varepsilon) \right|^p d\varepsilon \qquad (12)$$

While computationally expensive to evaluate, this will allow us to readily compare the proposed metrics to the true Wasserstein distance between rational spectra of normalized energy.[1]

The Wasserstein distance between two discrete measures in one dimension, such as discretized spectra, is even simpler to compute. A single pass over the data is sufficient if the measures are sorted (Peyré and Cuturi, 2018). In the numerical evaluation, we will make use of this to calculate $W$ between two spectra estimated using a traditional spectral-estimation technique, the Welch method.

## 2.2 Rational Spectra

A rational spectrum $\Phi(\omega)$ is a function of frequency $\omega$ that has a spectral factorization $\Phi(\omega) = G^*(i\omega)G(i\omega) = |G(i\omega)|^2$ where $G$ is a rational function, i.e., a function on the form $G(s) = B(s)/A(s)$ where $A, B$ are polynomials in the complex Laplace-variable $s$. In this work, we will limit our view to all-pole systems with $B(s) = \sigma \in \mathbb{R}^+$ and only simple poles, and will further assume, unless otherwise noted, that spectra are normalized to have unit energy, i.e.,

$$\int_{-\infty}^{\infty} \Phi(\omega) d\omega = \int_{-\infty}^{\infty} |G(i\omega)|^2 d\omega = 1$$

This allows us to calculate the optimal transport between two spectra.[2] All pole-systems are known to be accurate representations of any rational spectrum provided that the order is sufficiently high (Stoica and Moses, 2005), so this restriction is typically not limiting.[3]

We can expand $\Phi(\omega)$ by factorizing $A(s)$ according to

$$\Phi(s) = \frac{1}{A^*(s)A(s)} = \frac{1}{\prod_i (s - p_i)^*(s - p_i)} \qquad (13)$$

where $p$ denotes a root of $A$ (a pole of $G$) and $\cdot^*$ denotes the complex conjugate. Upon inspection of (13), it becomes clear that $\Phi(\omega)$ will be large for frequencies where $s = i\omega$ is close to a pole. This effect will be larger the closer the pole is to the imaginary axis, i.e., a pole with a large negative real part will never cause the denominator in (13) to be large, whereas a pole $p$ close to the imaginary axis will cause $\Phi$ to be very large when $i\omega = \Im p$.

A depiction of two rational spectra is shown in Fig. 2a. The locations of the poles of the spectra in the complex plane are shown in Fig. 2b. If we start at the origin and walk along the imaginary axis, we first encounter a pole in the blue spectrum, this corresponds to the first peak in the blue spectrum. We also notice that both spectra have poles appearing in complex conjugated pairs. This will always be the case for rational spectra with real coefficients, unless some poles fall on the real line.

Already now, we can intuitively feel that the distance between the poles of the two systems may be related to the transport distance between their spectra, something we are now ready to introduce.

## 2.3 A New Family of Metrics Between Rational Spectra

We have now come to the point where it is time to introduce the proposed metrics. We will start from the most simple metric and gradually work towards better approximations of the true optimal-transport distance. This approach will not only provide a gradual introduction of new concepts, but the initial simple metrics will remain very useful in some applications.

---

[1] The cumulative spectrum functions are conveniently obtained by integrating the spectrum using a differential-equation integrator.

[2] Optimal transport between measures of different mass can also be considered (Peyré and Cuturi, 2018).

[3] Spectra with deep notches require a high model order if represented by and all-pole model, but the more commonly encountered situation with sharp peaks in the spectrum is well represented by a low-order model.

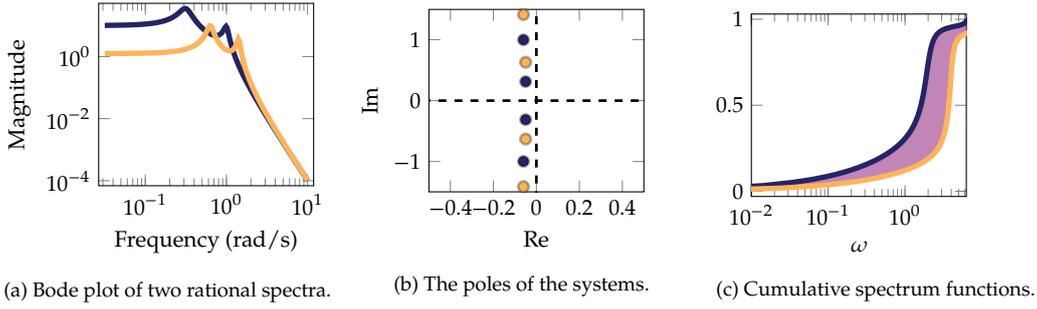

(a) Bode plot of two rational spectra.  (b) The poles of the systems.  (c) Cumulative spectrum functions.

Figure 2: Illustration of two rational spectra and the closed-form solution for the OT distance between them. The shaded area in the right figure corresponds to $W_1(\Phi_1, \Phi_2)$.

Guided by the intuition provided by Fig. 2, we propose the following distance between two linear time-invariant systems of the same order; For a particular ordering of the poles, let their root distance (RD) of order $p$ be defined as

$$\mathrm{RD}^p_p(\Phi_1, \Phi_2) = \sum_{i=1}^{n} |q_i - z_i|^p \tag{14}$$

where $n$ is the number of poles of the transfer functions and $q$ and $z$ denote the poles of the first and second systems respectively. This is simply the Euclidean distance between the poles of the two systems (to the power $p$), given a particular association between their poles. We will further discuss how to solve this association problem in general in Sec. 2.3.3, but it will turn out that it often suffices to simply sort the poles according to their imaginary value.

The root distance in its simplest form will turn out to be a crude approximation to the optimal-transport distance between two spectra. However, the vector-space property it implies will prove very useful, and this distance does not require the two spectra to be of equal energy, which is otherwise a limitation of the OT framework.

In a few simple situations, we can establish some relations between the RD and $W$. For single-pole systems and $p = 2$, this metric is actually equivalent to $W_2^2$, as evidenced by the following proposition:

**Proposition 1.** *Let $\Phi_1$ and $\Phi_2$ be two rational spectra with a single pole each with $\Re q = \Re z$, then*

$$\mathrm{RD}_2^2(\Phi_1, \Phi_2) = W_2^2(\Phi_1, \Phi_2) \tag{15}$$

*and an optimal transport map between them is given by*

$$T : i\omega \mapsto i\omega + q - z \tag{16}$$

*Proof.* The map $T$ trivially transports $\Phi_1$ to $\Phi_2$ by a distance $|q - z|$. Under some loose conditions, the optimal transport map when $p = 2$ is known to be the gradient of a convex function (Peyré and Cuturi, 2018). It thus suffices to provide such a function. In this case, it is given by

$$\psi(i\omega) = \frac{1}{2}(i\omega + q - z)^*(i\omega + q - z) + C \tag{17}$$

with $\nabla_{i\omega}\psi = T$ and $C$ an arbitrary constant. □

We remark that we can view $T$ in Proposition 1 as simply translating the imaginary axis according to the complex number $z - q$, or equivalently, transporting the spectrum along the imaginary axis by linear displacement. The equality constraint between the real parts of the poles is of course making the statement rather weak, but is required for the optimal transport problem to be well defined since in the single-pole setting, the energy (mass) of the spectrum is uniquely determined by the real part of the pole.

### 2.3.1 A note on normalization of the spectrum

One can envision two ways of performing normalization such that $F(\infty) = 1$. The correct way is to calculate the spectral energy of $G$ and scale $G$ accordingly. One could also attempt to normalize $F$ directly. These two methods are not equivalent, as $F$ is the integration of a squared quantity.

Before we develop the proposed metric further, we introduce another useful fact related to such scaling; scaling the poles of a transfer function has a predictable scaling effect on the energy in its spectrum:

**Proposition 2.** *Let $G$ be a rational transfer function, $\alpha$ denote a scalar, $n$ be the number of poles in $G$ and let $\tilde{G}$ denote a transfer function where all poles in $G$ have been scaled by $\alpha$. Then*

$$\alpha^{1-2n} \int \Phi(\omega) d\omega = \alpha^{1-2n} \int G^*(i\omega) G(i\omega) d\omega = \int \tilde{G}^*(i\omega) \tilde{G}(i\omega) d\omega \tag{18}$$

*Proof.* Simple calculations yield

$$\int \frac{1}{\prod_i (i\omega - \alpha p_i)^* (i\omega - \alpha p_i)} d\omega = \int \frac{\alpha^{-2n}}{\prod_i \left(i\frac{\omega}{\alpha} - p_i\right)^* \left(i\frac{\omega}{\alpha} - p_i\right)} d\omega = \tag{19}$$

$$= \int \frac{\alpha^{-2n}}{\prod_i (iu - p_i)^* (iu - p_i)} \alpha du = \alpha^{1-2n} \int G^*(i\omega) G(i\omega) d\omega \tag{20}$$

where the variable substitution $u = \omega/\alpha$ was done in the second step. $\square$

### 2.3.2 Partial Fractions and Residues of Rational Functions

One problem with the proposed root distance is that we pay an as high price for transporting poles with a high damping a certain distance as we do for transporting poles with low damping the same distance. The damping of a pole directly influences how much mass it contributes to the spectrum, and it would thus make sense to make it cheaper to transport highly damped poles, something we will make formal in this section. We will start by characterizing the spectral contribution of poles using tools from residue theory, and will use this mass to appropriately assign weights to the distances between individual poles.

Any strictly proper rational function with simple poles has a partial fraction decomposition on the form

$$G(s) = \sum_{i=1}^{n} \frac{r_i}{s - p_i}$$

where $r_i$ is the residue of $G$ at $p_i$ (Gamelin, 2003). There are several ways of calculating the residues, two simple ways are according to

$$r_i = \frac{B(p_j)}{A'(p_i)} = \frac{B(p_j)}{\prod_{j \neq i} p_i - p_j} \tag{21}$$

where $A'$ denotes the derivative of $A$.

The partial fraction decomposition of $G$ gives us a view in which the power in the spectrum is a linear combination of contributions from each pole, weighted by the pole's residue. Granted, this residue is a function of all the other poles, but if we neglect that fact for a while, we could imagine weighting the terms in (14) with the residues of the poles. We thus define the weighted root distance (WRD) by

$$\text{WRD}_p^p(\Phi_1, \Phi_2) = \sum_{i=1}^{n} (w_i^q w_i^z)^\beta \left| w_i^q q_i - w_i^z z_i \right|^p \tag{22}$$

$$\beta = \frac{1-p}{2} \tag{23}$$

$$w_i^p = \frac{-\pi |r_i|^2}{\Re p_i} = \int_{-\infty}^{\infty} \left| \frac{r_i}{i\omega - p_i} \right|^2 d\omega \tag{24}$$

where $w_i^p$ is termed the residue weight associated with pole $p_i$. We now proceed to provide some further indications that this weighting is reasonable, in particular the presence of the mysterious exponent $\beta$, in the form of a series of propositions.

**Proposition 3.** *When the poles of a rational spectrum $\Phi$ are scaled by $\alpha \in \mathbb{R}^+$, each residue is scaled by $\alpha^{1-n}$ and each residue weight is scaled by $\alpha^{1-2n}$.*

*Proof.* The proof is simple and follows from (21). The power $(1-n)$ comes from the product over all $i \neq j$. □

**Corollary 1.** *When the poles of a rational spectrum $\Phi$ are scaled by $\alpha \in \mathbb{R}^+$, the weighted root distance is scaled by $\alpha^{1-2n+p}$.*

*Proof.* Proposition 3 directly leads to

$$\sum_{i=1}^n \alpha^{2\beta(1-2n)} (w_i^q w_i^z)^\beta \left| \alpha^{2-2n}(w_i^q q_i - w_i^z z_i) \right|^p = \alpha^{2\beta(1-2n)+p(2-2n)} \sum_{i=1}^n (w_i^q w_i^z)^\beta \left| w_i^q q_i - w_i^z z_i \right|^p \quad (25)$$

For $\beta$ given by (23), the scaling factor becomes $\alpha^{1-2n+p}$. □

Corollary 1 provides us with a rate of growth of the WRD as a function of a scaling of the energy of the spectra between which the distance is calculated. The following proposition provides a similar result for the Wasserstein distance:

**Proposition 4.** *When the poles of a rational spectrum $\Phi$ are scaled by $\alpha \in \mathbb{R}^+$, $W_p^p$ is scaled by $\alpha^{1-2n+p}$.*

*Proof.* Let $\tilde{\Phi}$ denote a spectrum where poles are scaled by $\alpha$. From the proof of Proposition 2 we see that we may interpret the scaling of the poles with $\alpha$ as a scaling of the numerator with $\alpha^{-2n}$ together with a scaling of the frequency axis where the spectrum is evaluated by $1/\alpha$. We thus have

$$W_p^p(\tilde{\Phi}_1, \tilde{\Phi}_2) = \min_\gamma \int |\omega_1 - \omega_2|^p \, d\gamma\left(\frac{\omega_1}{\alpha}, \frac{\omega_2}{\alpha}\right) \quad (26)$$

$$= \min_\gamma \int \int \left| \alpha \frac{\omega_1}{\alpha} - \alpha \frac{\omega_2}{\alpha} \right|^p \gamma\left(\frac{\omega_1}{\alpha}, \frac{\omega_2}{\alpha}\right) d\omega_1 d\omega_2 \quad (27)$$

which after a change of variables $u_i = \omega_i/\alpha$ becomes

$$\alpha^{2+p} \min_\gamma \int \int |u_1 - u_2|^p \, \gamma(u_1, u_2) du_1 du_2 \quad (28)$$

The constraints take the form

$$\int \gamma\left(\frac{\omega_1}{\alpha}, \frac{\omega_2}{\alpha}\right) d\omega_2 = \alpha^{-2n} \Phi_1\left(\frac{\omega_1}{\alpha}\right) \quad (29)$$

$$\int \alpha^{2n+1} \gamma(u_1, u_2) du_2 = \Phi_1(u_1) \quad (30)$$

The $\alpha^{2n+1}$ factor can be moved to the objective function by rescaling $\gamma = \alpha^{-2n-1} \tilde{\gamma}$

$$W_p^p(\tilde{\Phi}_1, \tilde{\Phi}_2) = \alpha^{1-2n+p} \min_{\tilde{\gamma}} \int |u_1 - u_2|^p \, d\tilde{\gamma}(u_1, u_2) \quad (31)$$

$$\text{s.t.} \int \tilde{\gamma}(u_1, u_2) du_2 = \Phi_1(u_1), \quad \int \tilde{\gamma}(u_1, u_2) du_1 = \Phi_2(u_2) \quad (32)$$

□

These propositions together mean that both $W$ and the WRD have the same rate of groth when roots are scaled. This also means that the two distances have the same rate of growth when the mass of the spectra changes.

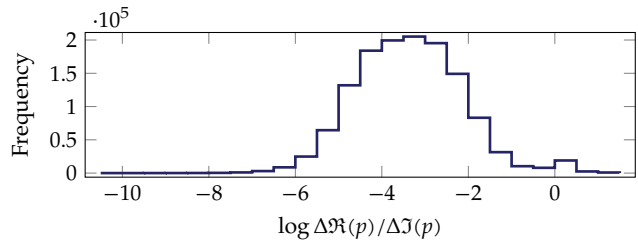

Figure 3: Ratios of separation between poles along the real axis and imaginary axis. Poles come from models of order 30, estimated from over 130 000 environmental sound recordings. The majority of poles are close to the imaginary axis in relation to how close the are to another pole. Note that the logarithmic scale on the horizontal axis might misrepresent the integral of this curve, approximately 98 % of the poles have a ratio of below 0.5.

### 2.3.3 The Assignment Problem

The root-based distances proposed thus far assume that there exists a meaningful assignment between the poles, i.e., each pole in one spectrum is compared to a single pole in the other. In general, poles should be matched based on their similarity, but what defines similarity between poles may not be obvious. We argue that the assignment that minimizes the root distance is a meaningful choice. Finding this assignment is itself an instance of an optimal-transport problem, often solved using specialized algorithms such as the Hungarian algorithm (Peyré and Cuturi, 2018).

In many applications, however, the real component of poles is very small in relation to the separation between the imaginary components of poles. In this case, the assignment implied by simply sorting the poles based on their imaginary value has a very high probability of optimally solving the assignment problem. Figure 3 illustrates a histogram of the ratio between separation in real and imaginary parts of poles in over 130 000 rational spectra of order 30, estimated from environmental acoustic recordings. This strategy has the added benefit that the assignment is uniquely determined by looking at a single system in isolation. The sorting can thus be done once per system, even if many pair-wise distances between spectra are to be calculated. All numerical evaluations in this paper uses this simple sorting heuristic.

While the sorting-based heuristic presented in this section is useful, we would also like to point out some pathological instances where it performs poorly; When the real part differs greatly between two poles with similar imaginary part. A large spectral mass in one spectrum might then be associated with the wrong pole in the other, making the distance much larger than the corresponding optimal-transport distance.

In the next section, we will introduce a distance which does not suffer from this limitation and is indeed a true transport-based distance.

### 2.3.4 Roots With Different Masses

In the previous section, we attributed a weight to each pole. The distance between two poles with different weights was assigned a weight corresponding to the geometric mean of the weights of the poles. From an OT perspective, this corresponds to a relaxation of the OT problem, typically referred to as unequal mass transport (Peyré and Cuturi, 2018). With this view, we allow ourselves to modify some masses in order make the problem simpler. We do not have to simplify the problem this much, though, as we will see in this section. If we let the previously defined residue weights correspond to masses, we can solve the OT problem between weighted poles directly. This way, the mass of a heavy pole can be distributed to several nearby lighter poles in the other spectrum so as to keep the total mass constant.

We set out motivating this work partly by the computational expense of solving OT problems between spectra, and we may therefore ask ourselves if we have gained anything by arriving back at solving an OT problem. Solving the OT problem between the poles is comparatively cheap, as the number of poles required to represent a spectrum is small, much smaller than the number of frequency bins required to accurately discretize the spectrum.

The OT problem between two sets of discrete masses can be solved very efficiently using the Sinkhorn algorithm (Peyré and Cuturi, 2018), or, since the measures involved are supported on a comparatively small number of points, using linear programming. We will refer to the distance defined by computing the optimal-transport distance between poles, with masses defined by the residue weights, as the Optimal-transport root distance (OTRD), the last distance we will introduce and also that which is closest to the true Wasserstein distance between spectra as will be evidenced in the following sections. Formally, the OTRD is defined as follows

$$\text{OTRD}_p^p(\Phi_1, \Phi_2) = \min_\gamma \sum \gamma \odot D$$
$$\text{s.t.} \sum_j \gamma[i,j] = w_1[i], \quad \sum_i \gamma[i,j] = w_2[j]$$

where $D \in \mathbb{R}^{n \times n}$ is the matrix containing all pairwise Euclidean distances, raised to the order $p$, between the $n$ poles in $\Phi_1$ and those in $\Phi_2$, and $w$ is defined by (24). Note, that this distance can be computed also for rational spectra with unequal number of poles, in which case $\gamma$ and $D$ will be non-square matrices. It can also be computed between spectra of unequal mass by means of unbalanced mass transport (Séjourné et al., 2019).

### 2.3.5 Unbalanced transport

There are situations in which one would like to avoid fully transporting all mass between two spectra. A few such cases are 1) The two spectra do not have the same mass. In this case, the standard, balanced, optimal-transport problem is unfeasible. 2) Energy is somehow lost or added to one spectrum in a way that should not be accounted for by transport. This could be the case if spectral energy is absorbed by a channel through which a signal is propagated—In this case it might not be desirable to transport mass from the other spectrum away from the absorbed (dampended) frequency. If, on the other hand, spectral energy is added to one spectrum by a noise source, this energy should ideally not be considered for transport and should rather be destroyed.

In Sec. 4.1 we demonstrate the use of unbalanced optimal root transport when one input spectrum contains a strong noise component.

## 2.4 Qualitative Behaviour of Distances

In this section, we will investigate how the different distances behave as we vary some aspect of two signals to be compared. In each experiment, two signals are generated and rational spectra are estimated by means of the least-squares method (Stoica and Moses, 2005). We will compare four distances, $W_2$ denotes the closed form solution (12) for $p = 2$, $W_2$Welch denotes Wasserstein distance between discrete spectra estimated by the Welch method, WRD and OTRD denote the proposed weighted and Optimal-transport root distances respectively.

In the first experiment, we compare two sinusoids. Figure 4 illustrates the different distances as functions of the frequency of one of the signals. In this example, the WRD and OTRD corresponds exactly to $W_2$. Calculating $W_2$ between Welch periodograms on windows of length 128 yields a result which on a large scale agrees with the model-based distances, but has a very noisy gradient.

Figure 5 illustrates an experiment where the compared signals are generated by filtering Gaussian white noise through filters with varying cutoff frequencies. The figure illustrates that the two root-based distances behave qualitatively similar to $W_2$.

We also investigate how the distances behave when measuring the distance between pairs of spectra estimated from environmental acoustic recordings. In Fig. 6, we estimate rational spectra of order 20 for 100 pairs of recordings containing bird calls. The figure indicates that the OTRD is a very good approximation of the true Wasserstein distance over the whole range of distances between signals. The WRD is a cruder approximation, but for $p = 1$ it remains a very reasonable approximation.

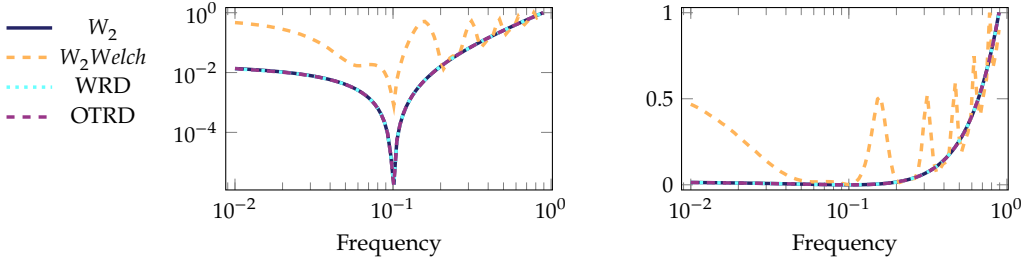

Figure 4: A signal with fixed frequency $f = 0.1$ is compared to $s(t) = \sin(2\pi f t)$ for varying $f$. A $2^{\text{nd}}$-order model is fit to both signals. The left and right plots show the same information with different scaling on the $y$-axis. All lines are re-scaled to have maximum value 1.

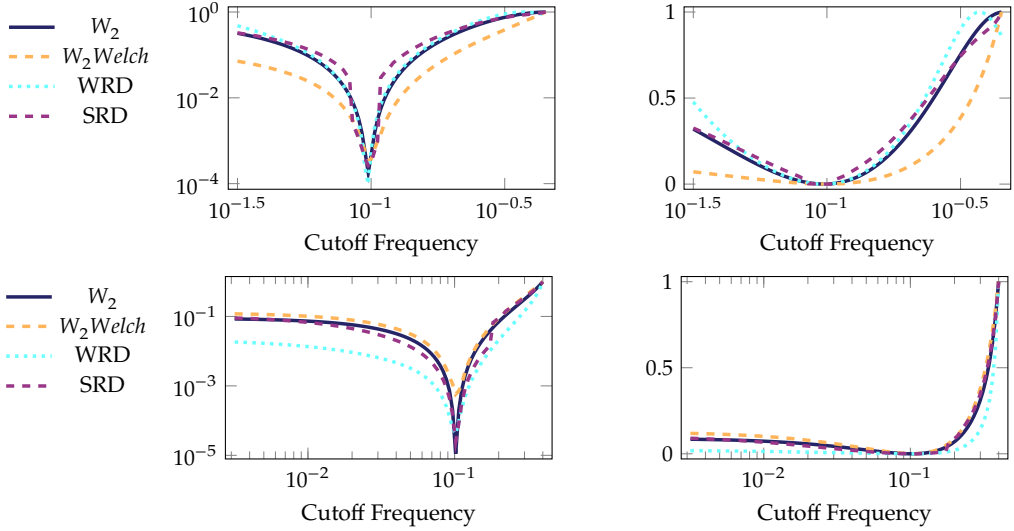

Figure 5: The compared signals are Gaussian white noise filtered through a lowpass (top)/ highpass (bottom) filter with varying cutoff frequency. The figures shows the distance between $s_1(t)$ and $s_2(t)$, where the cutoff frequency for $s_2$ varies. In both cases, the cutoff frequency for $s_1(t)$ was 0.1 and a $6^{\text{th}}$ order model was fit.

## 3 Barycenters and Displacement Interpolation Between Spectra

Some distances imply the existence of a shortest path, a geodesic. An interpolation is essentially a datapoint on that shortest path. Direct interpolation between two spectra under the $L_2$ norm causes a smooth blending in intensity between the two spectra, where peaks in one slowly fade as peaks in the other grow. In some situations, it is more desirable for frequencies in one spectra to slowly transition into frequencies in the other, i.e., an interpolation in pitch. This kind of interpolation, commonly referred to as displacement interpolation (Peyré and Cuturi, 2018), is exactly what corresponds to a geodesic (shortest path) between two spectra under the $W_2$ distance. In one dimension, this interpolation can be performed in closed form using inverses of cumulative spectrum functions. However, under the (weighted) root distance, displacement interpolation reduces to standard linear interpolation. This follows immediately from the fact that the root distance is equal to the Euclidean distance between roots. Furthermore, the computation of barycenters reduces to the standard arithmetic average of the poles of the systems, while calculation of the barycenter under $W_2$ involves a large optimization problem (Schmitz et al., 2018).

The fact that the metric space is Euclidean is immensely useful, it implies that standard techniques such as the K-means clustering algorithm and linear low-rank models such as PCA, are meaningful and apply directly.

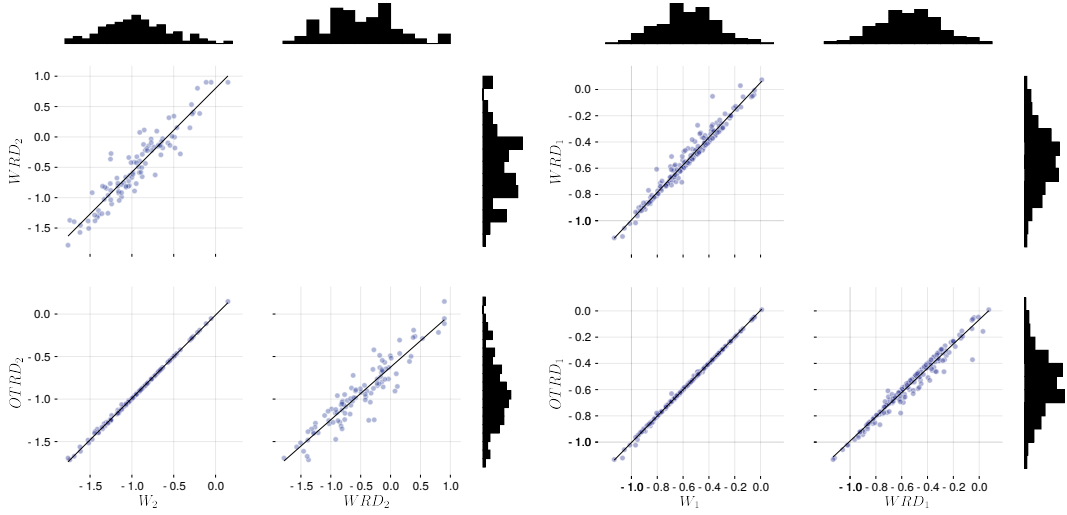

Figure 6: Correlation between log-metrics for 100 pairs of environmental audio recordings containing bird calls. On the left $p = 2$ and on the right with $p = 1$. The Optimal-transport root distance is a very good approximation to the true Wasserstein distance. The Weighted root distance is a slightly worse approximation for $p = 2$, but is faster to evaluate and is not limited to spectra of equal mass.

We also note that this works between spectra of different mass, something that requires extra attention in order to handle properly under the $W_2$ metric. To illustrate this property, we randomly generated 50 passbands for a bandpass filter and then generated 50 signals by filtering random noise inputs of length 500 through each of the filters. We then fit both rational spectra of order 6, and Welch spectra on windows of length 128, to each of the 2500 signals, and projected the roots and the Welch spectra respectively onto their first two principal components. The result is shown in Fig. 7, which illustrates how root embeddings from the same system are clustered tightly together in the $L_2$ sense, whereas the same is not true for the Welch spectra.

For the weighted root distance, we note that each individual pole has a weight. A weighted average of pole $j$

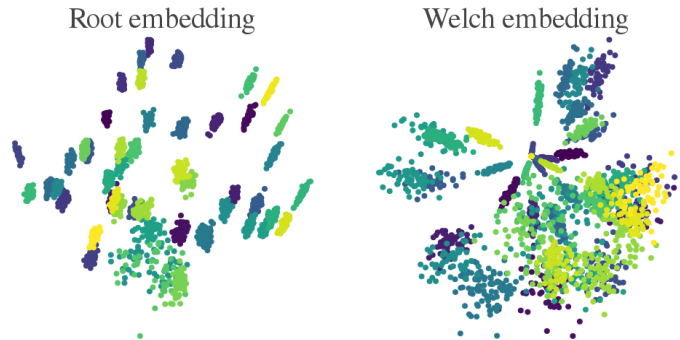

Figure 7: The first two principal components of roots to the left, Welch spectra to the right. The data consists of 2500 signals generated by 50 different bandpass systems.

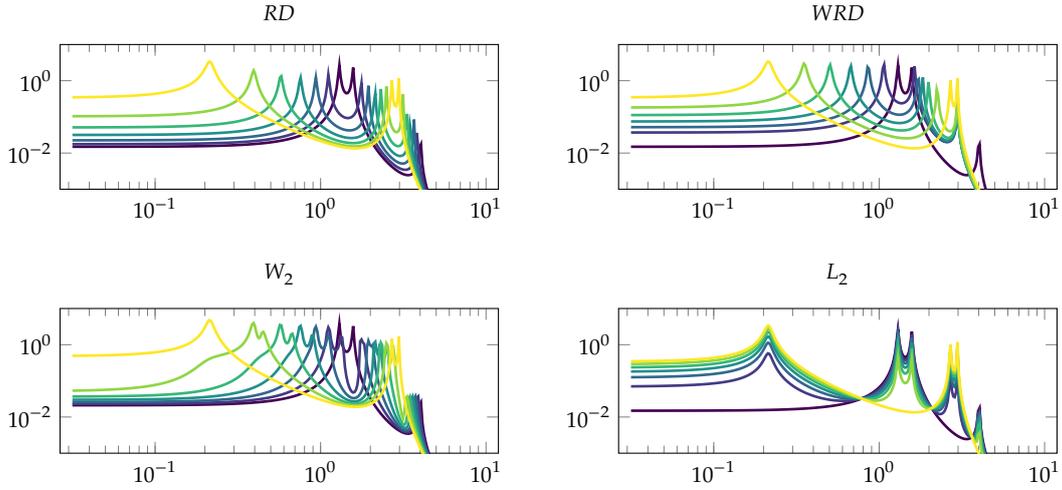

Figure 8: Direct interpolation under the $L_2$ norm and displacement interpolation under transport-based distances. Interpolation under the $L_2$ norm exhibits progressively fading peaks in one spectrum while peaks in the other grows. Interpolation under the transport-based distances instead displaces energy by shifting peaks along the frequency axis. This figure highlights how under $W_2$, energy from two different peaks in the blue spectrum is transported to the same peak in the yellow spectrum, whereas for the root-based distances, each peak in the blue spectrum is associated with a single peak in the yellow spectrum.

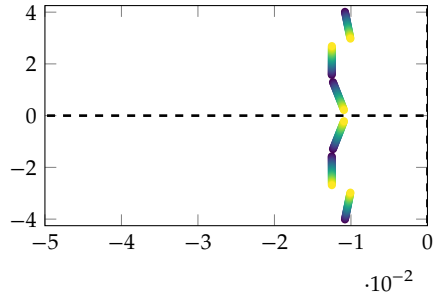

Figure 9: Illustration of how the poles move in the complex plane while interpolating under the root distance. The poles trace the path of the geodesic, and this figure serves as an illustration of how the distance is measured between the poles.

for $N$ systems thus takes the form

$$\mu_j = \frac{\sum_{i=1}^{N} w_{ij} p_{ij}}{\sum_{i=1}^{N} w_{ij}}$$

Examples of interpolation under different metrics are shown in Fig. 8 and Fig. 9 illustrates how the poles move in the complex plane while interpolating under the root distance, corresponding to the upper left plot in Fig. 8.

One possible extension of the concept of an interpolation to more than two points is that of a weighted *barycenter*. Formally, a barycenter $Q$ with barycentric coordinates $\lambda$, of a set of measures $\{G_i\}$, is defined as

$$\arg\min_{Q} \sum_i \lambda_i W(G_i, Q) \tag{33}$$

where $W$ is the metric. This problem comes in different flavors depending on the metric and on which aspects of $Q$ are being optimized. If only the weights of $Q$ are being optimized, the problem is convex (Cuturi and Doucet, 2014). If also the support of $Q$ (for the OTRD, the support is the set of pole locations) is optimized, the problem is non-convex, but can be solved for a local minimizer.

The calculations of barycenters is central to many unsupervised clustering algorithms, something that will be explored in Sec. 4.2.2, where we use the algorithm by Cuturi and Doucet (2014) to calculate barycenters.

## 3.1 Barycentric Coordinates

The inverse problem to that of finding a barycenter is that of finding the barycentric coordinates $\lambda$ of a query point $Q$, such that the resulting barycenter is as close as possible to the query point. Given a set of rational spectra $\{G_i\}$, a nonlinear projection of a spectrum $Q$ onto this set can be obtained by solving the following nested optimization problem

$$\lambda = \arg\min_{\bar{\lambda}} \ W(Q, Q^*(\bar{\lambda})) \quad (34)$$

$$Q^*(\bar{\lambda}) = \arg\min_{\bar{Q}} \sum_i \bar{\lambda}_i W(G_i, \bar{Q}) \quad (35)$$

where $\lambda$ are the barycentric coordinates belonging to the probability simplex. Problems of this type are sometimes referred to as *histogram regression* (Bonneel, Peyré, and Cuturi, 2016).

A nonlinear projection onto a basis consisting of spectra can be useful for, e.g., spectral dictionary learning, basis pursuit, topic modelling, denoising and detection. We will illustrate an example of this kind of projection in Sec. 4.1 in an application of detection of marine vessels.

# 4 Applications

In this section, we provide examples and insight into some applications of the proposed metrics.

## 4.1 Detection and Tracking of Marine Traffic

Underwater sensors are commonly deployed in order to track the movement of marine vessels in near-shore conditions, where each vessel has a unique, velocity dependent, frequency signature (Chung et al., 2011; Vishnu and Chitre, 2017). We demonstrate two potential applications.

Given a composite spectrum created by the cavitation noise of several vessels, and a database/dictionary of spectral signatures for different vessels, one may decompose an observed signal by projecting it onto the dictionary through the calculation of the barycentric coordinates. Non-zero barycentric coordinates indicate the presence of a vessel. For the OTRD, the dictionary and query spectra need not have the same number of poles, which is utilized in the example below.

Successful tracking of vessels using their noise spectra requires pairing of the spectra observed by one sensor with spectra observed by another sensor. Underwater communication bandwidth is notoriously low, necessitating a compact representation of the information observed by one sensor to be transmitted for pairing purposes. In a situation like this, a sensor can transmit a small number of linear model coefficients, which can be used by the receiver to derive the spectrum of the signal observed by the sender as well as for finding the closest matching spectrum observed by the receiver.

In Fig. 10, we illustrate the usage of barycentric-coordinate projection for ship detection using estimated DEMON spectra.[4]

We remark that in the absence of Doppler effects, the LASSO estimator on regular spectral estimates would do a very good job at decomposing a spectrum as a sum of dictionary spectra. However, in the presence of even slight frequency shifts, such an approach breaks down due to the typically very narrow peaks in DEMON spectra.

To demonstrate an application of unbalanced mass transport, we add an additional noise component to the spectrum of the mixed signal, shown to the right in Fig. 11 where the noise component shows up as the third peak in the mixed spectrum. We then calculate the transport plan between the mixed signal and all the three dictionary signals, while penalizing the KL divergence between the transport map marginals and the weights of the spectra (Séjourné et al., 2019). The resulting transport plan, $\gamma$, is shown on the right in Fig. 11, where we see that very little mass was transported to the third (and by conjugate symmetry the eighth) pole of the mixed signal, and almost no mass at all was transported from the third signal. In this example, we assigned a unit mass to each pole in order to make the transport map easy to interpret.

---

[4] A DEMON spectrum (Detection of Envelope Modulation on Noise) (Chung et al., 2011; Vishnu and Chitre, 2017) is commonly employed to extract a spectrum in which modulation frequencies are easy to identify.

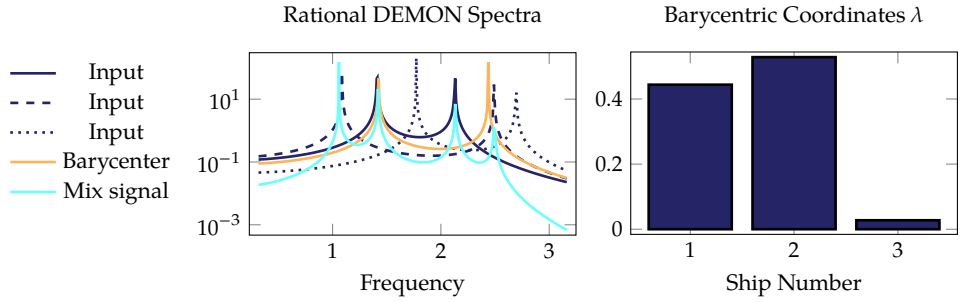

Figure 10: The left panel shows estimated rational DEMON spectra of three marine vessels in blue. Noise from the first two ships were superimposed, from which the rational DEMON spectra labeled *Mix signal* of twice the order was estimated. The barycenter of all three input spectra is also shown. In the right panel, the barycentric coordinates of the mix spectrum are shown, indicating that the nonlinear projection indeed identifies the two spectra contributing to the mixed signal.

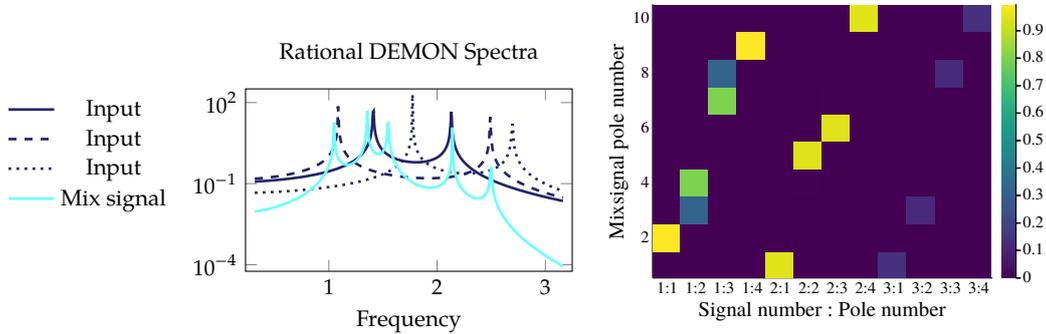

Figure 11: Unbalanced mass transport between three spectra and a mixed spectrum containing the two first input spectra and an additional noise component, appearing as the third peak in the mixed spectrum. In the optimal transport plan $\gamma$ to the right, very little mass has been transported to the noise component, and almost no mass has been transported from the third input spectrum, which did not contribute to the mixed spectrum.

### 4.2 Classification and Clustering of Signals

We now turn to a classical machine-learning application, supervised classification and unsupervised clustering of signals. We consider a training dataset with $N = 21500$ audio clips of approximately 0.6 s length containing bird calls from 30 distinct species of birds.

#### 4.2.1 Supervised Learning

For supervised classification, the root distance (RD) was used to find the nearest neighbor within the training set for each audio clip in a test set of 3037 samples, an operation that is very fast due to the Euclidean nature of the RD lending itself to acceleration by the use of a KD-tree. In 97.3% of the cases, the nearest neighbor was from the same class, indicating that the simple root distance together with a simple nearest-neighbor classifier can produce meaningful results for a very small computational cost. This further implies that density-based clustering using the proposed metrics is likely to produce clusters coherent with the species of bird present. All of this was without considering a time-frequency representation, but rather by naively assuming a stationary signal and estimating a single spectrum.

If the same exercise is carried out by extracting the height and location of the peaks of spectra estimated using Welch's method on the same dataset, the nearest-neighbor classifier obtains a 86.9% accuracy. The same number of peaks and locations as the number of poles in the rational spectra were used.[5]

### 4.2.2 Unsupervised Learning

We demonstrated above the use of the root distance for supervised learning. Akin to the K-means clustering algorithm, one may adopt a similar approach using a transport-based distance and arrive at the K-barycenter clustering algorithm (Cuturi and Doucet, 2014) for unsupervised learning. Similar to K-means, one iterates between assigning data points to the closest barycenter under the chosen metric, and finding new barycenters based on the cluster assignments. This method, while on the surface having similar complexity as the K-means algorithm, is substantially more expensive due to the K-means algorithm being accelerated by the use of data structures with $\mathcal{O}(\log(N))$ lookup, such as a KD-tree, for nearest neighbor queries. Nevertheless, we showcase this algorithm on a reduced version of the same dataset of bird calls.

The K-barycenter algorithm with the optimal-transport root distance was run on 20 samples from each class of bird calls. Figure 12 illustrates the resulting cluster assignments. Interestingly, if cluster assignments are matched with the best matching class of bird, the corresponding supervised accuracy was 41%, indicating that the clusters are meaningful. The accuracy obtained by random assignment would, on average, be 1/30 = 3.3%. We emphasize that this algorithm is not designed to perform classification, but it nevertheless found clusters largely agreeing with the species of bird present.

The clustering took 10s to compute on a laptop for this dataset with 600 LTI systems and 30 barycenters using 10 iterations in the K-barycenter algorithm. While computationally more expensive than the K-means algorithm under the root distance, this algorithm with its $\mathcal{O}(NK)$ cost is substantially faster than clustering approaches that require the computation of all pairwise distances, which have an $\mathcal{O}(N^2)$ cost.

## 4.3 Gradient-Based Learning

The proposed distances are all differentiable with respect to the poles of the rational spectrum. If the spectrum is estimated in a way that makes the poles differentiable with respect to the signal, it is possible to backpropagate through these distances and to use them as loss functions for training of, e.g., autoencoders. Since the rational spectra can be readily estimated using either the SVD or the normal equations, the estimation does not pose any problems for differentiation. Calculating the poles of a rational spectrum is equivalent of an eigenvalue calculation, which is also differentiable almost everywhere.

Further, preconditioners derived from the distances have shown promise for gradient-based optimization of LTI controllers and system identification, something we intend to explore in detail in upcoming work.

## 5 Discussion

The three metrics proposed vary in their levels of complexity, as well as in their levels of approximation to the true optimal-transport distance between rational spectra. Besides providing insight into the connection between spectra and optimal transport, the very compact representation of a rational spectrum, the poles of a rational function, makes computation and information transmission very efficient.

---

[5] Poles are complex numbers, but are twice redundant due to symmetry. The number of free parameters in the two representations were thus the same.

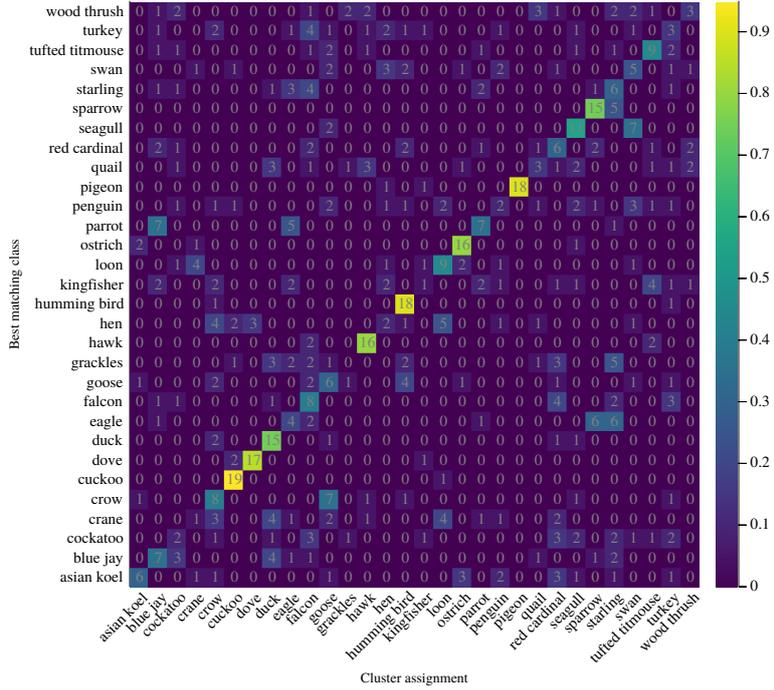

Figure 12: Unsupervised learning on 20 samples of bird calls from 30 different classes, clustered into 30 clusters. The figure shows the confusion matrix between cluster assignments and the best matching class of birds using the K-barycenter algorithm under the optimal-transport root distance. Interpreted as a supervised task, the class discrimination accuracy was 41%, compared to the random baseline of 1/30 = 3.3%.

Given a signal, the computation of a transfer function that approximates the spectrum of the signal is straightforward and can be done on closed form using, e.g., the least-squares method. The computational complexity of this is roughly $\mathcal{O}(Tn + n^{2.4})$ using a well-optimized BLAS library, where $n$ is the number of free parameters in the transfer function and $T$ is the length of the signal. For $n = 20$, a rational spectrum with $n/2 = 10$ peaks is obtained, which form a quite good approximation to, for instance, audio signals. In applications where $\mathcal{O}(N^2)$ pairwise distances between $N$ signals are needed, this computational cost is negligible, as it is carried out once per signal, or $N$ times, only.

The metric properties of the simple root distance also deserves some further discussion. The expression for $\text{RD}_2^2$, (14), can be equivalently written as

$$\left\| \begin{bmatrix} \Re Q \\ \Im Q \end{bmatrix} - \begin{bmatrix} \Re Z \\ \Im Z \end{bmatrix} \right\|^2$$

where the poles of each spectrum have been expanded into a vector consisting of the real parts stacked above the imaginary parts. It thus becomes clear that although the RD measures distances between measures defined on two-dimensional atoms (the poles), it can be interpreted as a simple $L^p$ distance between suitably chosen vectors. If we let $d$ denote such a vector, i.e., $d = [\Re Q^\mathsf{T} \, \Im Q^\mathsf{T}]^\mathsf{T}$, and $\mathcal{D}$ denote a dictionary matrix constructed of such vectors stacked horizontally, the barycentric coordinates of a query spectrum $Q$ with expanded root vector $q$ can be computed using the standard least-squares approach $\lambda = \arg \min \|\mathcal{D}\lambda - q\|$, s.t. $\sum \lambda = 1$, possibly with a sparsity-promoting regularization added. Parallels are easily drawn to standard techniques such as low-rank factorizations.

Avenues for future work include weighting schemes corresponding to working on log-magnitudes and log-frequency scales, optimal transport for low-order approximations of linear systems, preconditioning for gradient-based optimization of control systems, as well as methods for working with time-frequency representations, e.g., rational spectrograms.

## 6 Summary

We have defined three metrics between pairs of signals, systems or rational spectra, rooted in linear-systems theory, and made connections between them and the optimal-transport distance (Wasserstein distance). We demonstrated how these metrics can be used to solve machine-learning tasks such as unsupervised clustering and supervised classification of signals, and to perform detection and spectral matching in an application of marine-traffic tracking. We also illustrated how one of the metric spaces is Euclidean, which implies that displacement-interpolation between spectra reduces to linear interpolation and the calculation of barycenters reduces to taking a standard arithmetic average. This allowed immediate application of standard Euclidean techniques such as K-means for clustering and the use of KD-trees for efficient nearest-neighbor queries.

The metrics allow for efficient computation of distance or similarity between signals, displacement interpolation and barycenters of spectra under a transport-based metric and projection of spectra through calculation of barycentric coordinates.

Julia (Bezanson et al., 2017) implementations of all algorithms mentioned in this paper are provided under the MIT licence at github.com/baggepinnen/SpectralDistances.jl.